\documentclass[10pt,twocolumn,letterpaper]{article}

\usepackage{cvpr}
\usepackage{times}
\usepackage{epsfig}
\usepackage{graphicx}
\usepackage{amsmath}
\usepackage{amssymb}
\usepackage{helvet}
\usepackage{courier}
\usepackage{url}
\usepackage{multirow}
\usepackage{array}
\usepackage{color}

\usepackage[pagebackref=true,breaklinks=true,letterpaper=true,colorlinks,bookmarks=false]{hyperref}

\cvprfinalcopy 

\def\cvprPaperID{1994} 

\definecolor{CQColor}{rgb}{0.0,0.0,0.0} 
\newcommand{\cq}[1]{{\color{CQColor}#1}}
\def\OurNet{ADCrowdNet}

\ifcvprfinal\fi
\begin{document}
	
	\title{ADCrowdNet: An Attention-Injective Deformable Convolutional Network for Crowd Understanding}
	
	\author{Ning Liu$^{1, 2}$~~~~~Yongchao Long$^{1, 2}$~~~~~Changqing Zou$^3$~~~~~Qun Niu$^{1, 2}$~~~~~Li Pan$^4$~~~~~Hefeng Wu$^{1, 5, 6\,}$\thanks{\,The corresponding author is Hefeng Wu. This research is supported by the National Natural Science Foundation of China (Grant No. 91746204 and 61876045), and Opening Project of Guangdong Province Key Laboratory of Information Security Technology (Grant No. 2017B030314131).}
\vspace{1ex}\\
		$^1$School of Data and Computer Science, Sun Yat-sen University \\
		$^2$Guangdong Key Laboratory of Information Security Technology\;\;\; $^3$University of Maryland\\
		$^4$Shanghai Jiao Tong University \; $^5$Guangdong University of Foreign Studies \; $^6$WINNER Technology\\
		{\tt\small liuning2@mail.sysu.edu.cn, \{longych3, niuqun\}@mail2.sysu.edu.cn, }\\
		{\tt\small cqzou@umiacs.umd.edu, panli@sjtu.edu.cn, wuhefeng@gmail.com}
	}
	
	\maketitle
	\thispagestyle{empty}
	
	\begin{abstract}
	We propose an attention-injective deformable convolutional network called \OurNet{} for crowd understanding that can address the accuracy degradation problem of highly congested noisy scenes. \OurNet{} contains two concatenated networks. An attention-aware network called Attention Map Generator (AMG) first detects crowd regions in images and computes the congestion degree of these regions. Based on detected crowd regions and congestion priors, a multi-scale deformable network called Density Map Estimator (DME) then generates high-quality density maps. With the attention-aware training scheme and multi-scale deformable convolutional scheme, the proposed \OurNet{} achieves the capability of being more effective to capture the crowd features and more resistant to various noises. We have evaluated our method on four popular crowd counting datasets (ShanghaiTech, UCF\_CC\_50, WorldEXPO'10, and UCSD) and an extra vehicle counting dataset TRANCOS, and our approach beats existing state-of-the-art approaches on all of these datasets.
	\end{abstract}

 \section{Introduction}\label{sec:intro}
Crowd understanding has attracted much attention recently because of its wide range of applications like public safety, congestion avoidance, and flow analysis. The current research trend for crowd understanding has developed from counting the number of people to displaying distribution of crowd through density map. 
Generally, generating accurate crowd density maps and performing precise crowd counting for highly congested noisy scenes is challenging due to the complexity of crowd scenes caused by various factors including background noises, occlusions, and diversified crowd distributions. 

\begin{figure*}[]
	\centering
	\includegraphics[width=0.79\textwidth]{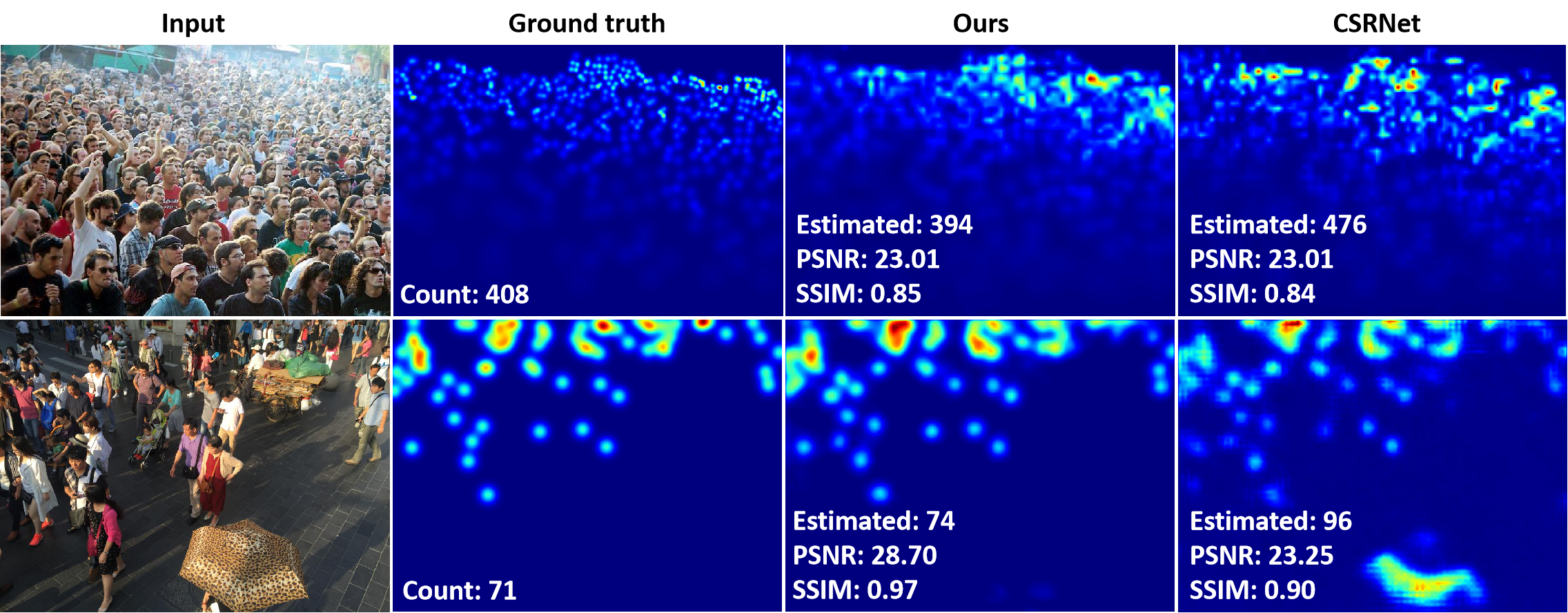}
	\caption{From left to right: a congested sample (top) and a noisy sample (bottom) from ShanghaiTech dataset~\cite{zhang2016single}, ground truth density map, and the generated density maps from the proposed \OurNet{} and the state-of-the-art method~\cite{li2018csrnet}. \OurNet{} outperforms the state-of-the-art method on both congested and noisy scenes.}
	\label{fig:sample}
	\vspace{-0.15in}
\end{figure*}

Researchers recently have leveraged deep neural networks (DNN) for accurate crowd density map generation and precise crowd counting. Although these DNNs-based methods~\cite{zhang2016single,sam2017switching,sindagi2017generating,liu2018crowd,CaoWZS18eccv,li2018csrnet} have made significant success in solving the above issues, they still have the problem of accuracy degradation when applied in highly congested noisy scenes. As shown in Figure~\ref{fig:sample}, the state-of-the-art approach~\cite{li2018csrnet}, which has achieved much lower Mean Absolute Error (MAE) than the previous state-of-the-art methods, is still severely affected by background noises, occlusions, and non-uniform crowd distributions. 

In this paper, we aim at an approach which is capable of dealing with highly congested noisy scenes for the crowd understanding problem. To achieve this, we designed an attention-injective deformable convolutional neural network called \OurNet{} which is empowered by a visual attention mechanism and a multi-scale deformable convolution scheme. The visual attention mechanism is delicately designed for alleviating the effects from various noises in the input. The multi-scale deformable convolution scheme is specially introduced for the congested environments. The basic principle of visual attention mechanism is to use the pertinent information rather than all available information in the input image to compute the neural response. This principle of focusing on specific parts of the input has been successfully applied in various deep learning models for images classification~\cite{hu2017squeeze}, semantic segmentation~\cite{ren2017end}, image deblurring~\cite{qian2018attentive}, and visual pose estimation~\cite{chu2017multi}, which also suits our problem where the interest regions containing the crowd need to be recognized and highlighted out from noisy scenes. The multi-scale deformable convolution scheme takes as input the information of the dynamic sampling locations, other than evenly distributed locations, which has the capability of modeling complex geometric transformation and diverse crowd distribution. This scheme fits well the nature of the distortion caused by the perspective view of the camera and diverse crowd distributions in real world, therefore guaranteeing more accurate crowd density maps for the congested scenes. 

\begin{figure*}[]
	\centering
	\includegraphics[width=0.99\textwidth]{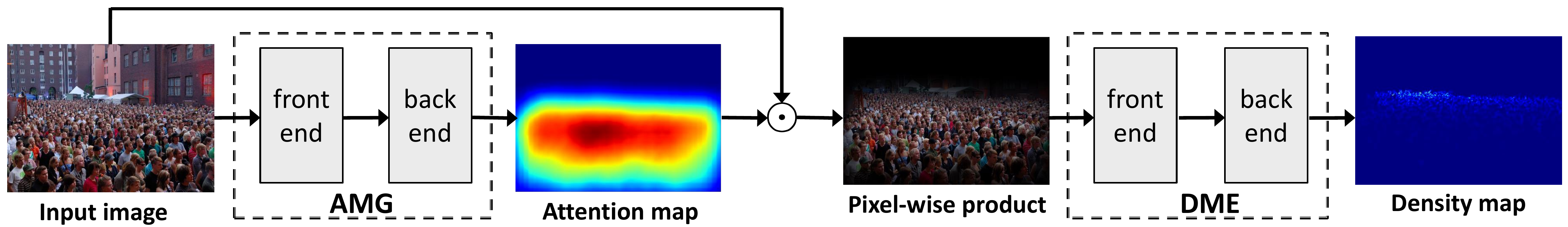}
	\caption{Architecture overview of \OurNet{}. The well trained AMG generates the attention map of the input image. The pixel-wise product of the input image and its attention map is taken as the input to train the DME network.}
	\vspace{-0.15in}
	\label{fig:overviewArch}
\end{figure*}

To incorporate the visual attention mechanism and deformable convolution scheme, we leverage an architecture consisting of two neural networks as shown in Figure~\ref{fig:overviewArch}. Our training contains two stages. The first stage generates an attention map for a target image via a network called Attention Map Generator (AMG). The second stage takes the output of AMG as input and generates the crowd density map via a network called Density Map Estimator (DME). 
The attention map generator AMG mainly provides two types of priors for the DME network: 1) candidate crowd regions and 2) the congestion degree of crowd regions. The former prior enables the multi-scale deformable convolution scheme empowered DME network to pay more attention to those regions having people crowds, and thus improving the capacity of being resistant to various noises. The latter prior indicates each crowd region with congestion degree (i.e., how crowded each crowd region is), which provides fine-grained congestion context
prior for the subsequent DME network and boosts the performance
of the DME network on the scenes containing diverse crowd distribution. 

The main contributions of this paper are summarized as follows. First, a novel attention-injective deformable convolutional network framework \OurNet{} is proposed for crowd understanding. Second, our AMG model that attends the crowd regions in images, is innovatively formulated as a binary classification network by introducing third party negative data (i.e., background images with no crowds). Third, our DME model can estimate the crowds effectively by using the proposed structure of aggregating multi-scale deformable convolution representations. Furthermore, extensive experiments conducted on all popular datasets demonstrate the superior performance of our approach over existing leading ones. In particular, the proposed model \OurNet{} outperforms the state-of-the-art crowd counting solution CSRNet~\cite{li2018csrnet}
with 3.0\%, 18.8\%, 3.0\%, 13.9\% and 5.1\% lower Mean Absolute Error (MAE) on ShanghaiTech Part\_A, Part\_B, UCF\_CC\_50, WorldExpo’10, UCSD datasets, respectively.
Apart from crowd counting, \OurNet{} is also general for other counting tasks. We have evaluated \OurNet{} on a popular vehicle counting dataset named TRANCOS \cite{guerrero2015extremely}, and \OurNet{} achieves 32.8\% lower MAE than CSRNet.  

	\section{Related Work}\label{sec:related}
\textbf{Counting by detection: }Early approaches of crowd understanding mostly focus on the number of people in crowds~\cite{DollarWSP12}. The major characteristics of these approaches are the sliding window based detection scheme and hand crafted features extracted from the whole human body or particular body parts  with low-level descriptors like Haar wavelets~\cite{viola2004robust} and HOG~\cite{dalal2005histograms}. Generally, approaches in these groups deliver accurate counts when their underlying assumptions are met but are not applicable in more challenging congested scenes. 

\textbf{Counting by regression: }Counting by regression approaches differs depending on the target of regression: object count~\cite{ChenLGX12,ChenGXL13}, or object density~\cite{lempitsky2010learning}. This group of approaches avoid solving the hard detection problem. Instead, they deploy regression model to learn the mapping between image characteristics (mainly histograms of lower level or middle level features) and object count or density. These approaches that directly regress the total object count discard the information of the location of the objects and only use 1-dimensional object count for learning. As a result, a large number of training images with the supplied counts are needed in training. Lempitsky \etal~\cite{lempitsky2010learning} propose a method to solve counting problem by modeling the crowd density at each pixel and cast the problem as that of estimating an image density whose integral over any image region gives the count of objects within that region. Since the ideal linear mapping is hard to obtain, Pham \etal~\cite{pham2015count} use random forest regression to learn a non-linear mapping instead of the linear one.


\textbf{Crowd understanding by CNN: }Inspired by the great success in visual classification and recognition, literature also focuses on the CNN-based approaches to predict crowd density map and count the number of crowds \cite{walach2016learning,onoro2016towards,LiHH2018arXiv,KangC18bmvc,qiu2019crowd}. 
Walach \etal~\cite{walach2016learning} use CNN with a layered training structure. Shang \etal\cite{shang2016end} adapt an end-to-end CNN which uses the entire images as input to learn the local and global count of the images and ultimately outputs the crowd count. A dual-column network combining shallow and deep layers is used in \cite{boominathan2016crowdnet} to generate density maps. In~\cite{zhang2016single}, a multi-column CNN is proposed to estimate density map by exacting features at different scales. Similar idea is used in \cite{onoro2016towards}. Marsden \etal~\cite{marsden2016fully} try a single-column fully convolutional network to generate density map while Sindagi \etal~\cite{sindagi2017cnn} present a CNN that uses high-level prior to boost accuracy.  

More recently, Sindagi \etal~\cite{sindagi2017generating} propose a multi-column CNN called CP-CNN that uses context at various levels to improve generate high-quality density maps. Li \etal~\cite{li2018csrnet} propose a model called CSRNet that uses dilated convolution to enlarge receptive fields and extract deeper features for boosting performance. These two approaches have achieved the state-of-the-art performances. 



\section{Attention-Injective Deformable Convolutional Network}\label{sec:detail}
The architecture of the proposed \OurNet{} method is illustrated in Figure~\ref{fig:overviewArch}. It employs two concatenated networks: AMG and DME. AMG is a classification network based on fully convolutional architecture for attention map generation, while DME is a multi-scale network based on deformable convolutional layers for density map generation. Before training DME, we train the AMG module with crowd images (positive training examples) and background images (negative training examples). We then use the well-trained AMG to generate the attention map of the input image. Afterward, we train the DME module using the pixel-wise product of input images and the corresponding attention maps. In the following sections, we will detail the architectures of the AMG and DME netwroks.

\begin{figure}[]
	\centering
	\includegraphics[ width=0.46\textwidth]{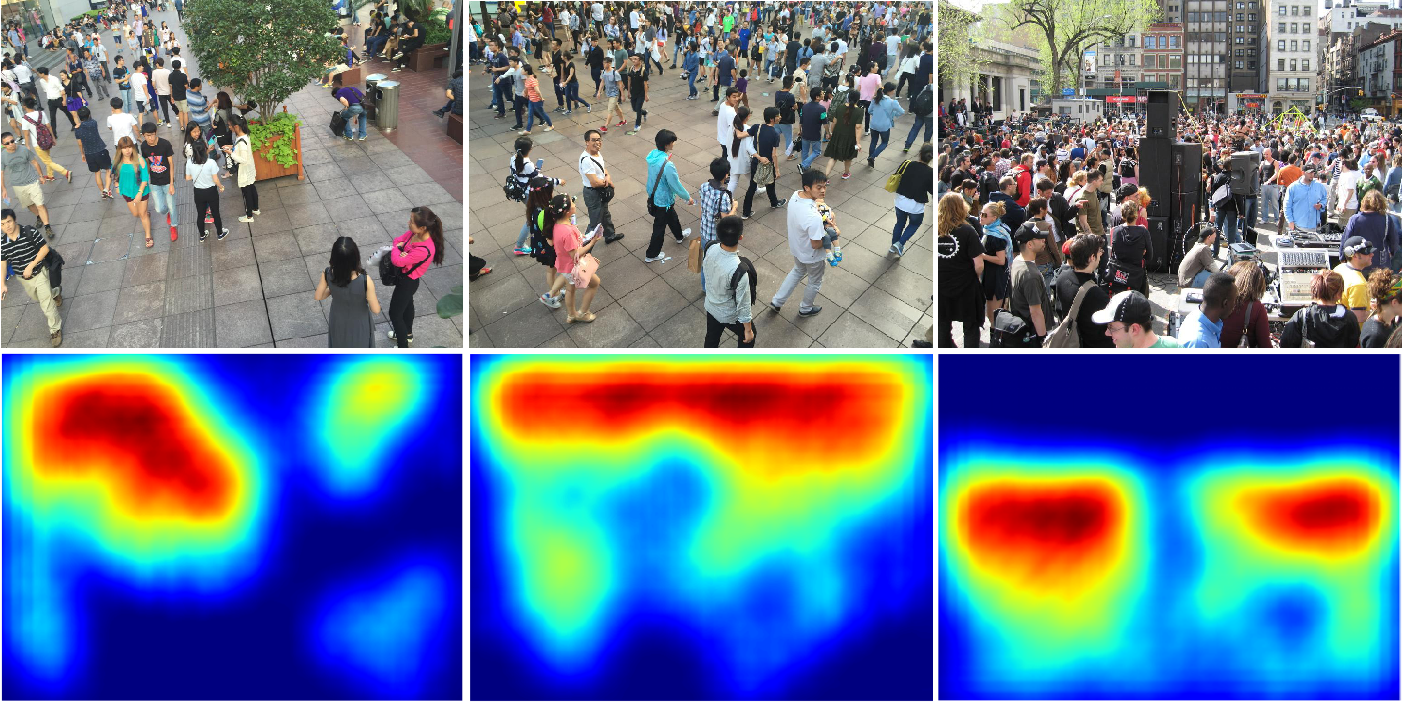}
	\caption{Attention maps generated by AMG at various crowd density levels (density level increases from left to right).}
	\vspace{-0.2in}
	\label{fig:ddlevel}
\end{figure}

\subsection{Attention Map Generator}
\subsubsection{Attention map}
Attention map is an image-sized weight map where crowd regions have higher values. In our work, attention map is a feature map from a two-category classification network AMG which classifies an input image into crowd image or background image.
The idea of using feature map to find the crowd regions in the input is motivated by an object localization work~\cite{zhou2016learning} which points out that the feature maps of classification network contain the location information of target objects.

The pipeline of the attention map generation is shown in Figure~\ref{fig:AMG}. $F_c$ and $F_b$ are the feature maps from the last convolution layer of AMG. $W_c$ and $W_b$ are the spatial average of the $F_c$ and $F_b$ after global average pooling (i.e., GAP in Figure~\ref{fig:AMG}). $P_c$ and $P_b$ are confidence scores of the predicted two class. They are generated by softmax from $W_c$ and $W_b$. The attention map is obtained by up-sampling the linear weighted fusion of the two feature maps $F_c$ and $F_b$ (i.e., $F_c \cdot P_c + F_b \cdot P_b$) to the same size as the input image. We also normalize the attention map such that all element values fall in the range $[0, 1]$.


The attention map highlights the regions of crowds. In addition, it also indicates the degree of congestion in individual regions, i.e., higher congestion degree values indicate more congested crowds and lower values indicate less congested ones. Figure~\ref{fig:ddlevel} illustrates the effect of attention maps at different density levels. The pixel-wise product between the attention map and the input image produces the
input data used by the DME network.

\subsubsection{Architecture of attention map generator}
\begin{figure}[]
	\centering
	\includegraphics[width=0.46\textwidth]{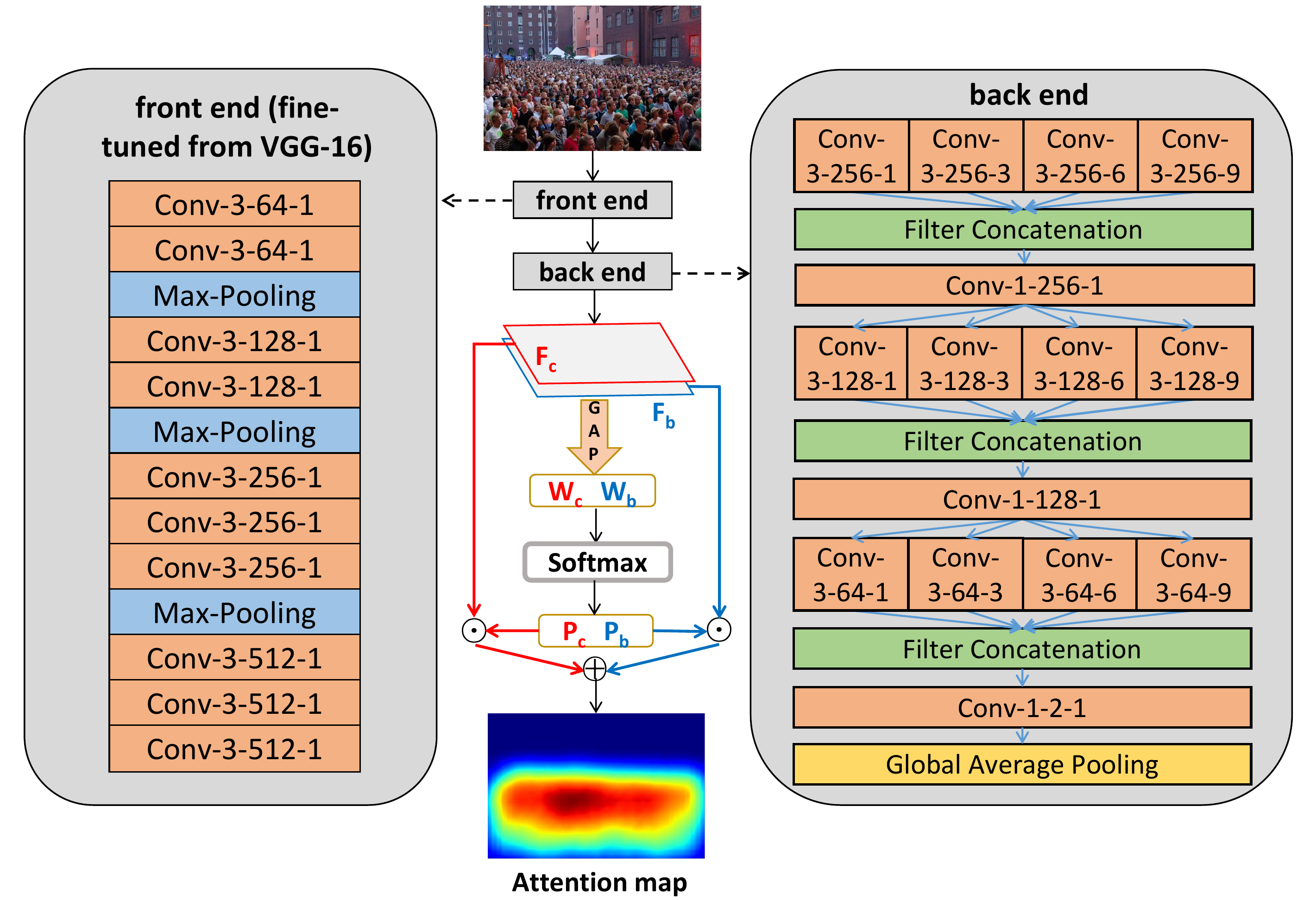}
	\caption{Architecture of AMG. All convolutional layers use padding to maintain the previous size. The convolutional layers' parameters are denoted as ``Conv-(kernel size)-(number of filters)-(dilation rate)'', max-pooling layers are conducted over a 2$\times$2 pixel window, with stride 2.}
	\vspace{-0.1in}
	\label{fig:AMG}
\end{figure}

The architecture of AMG is shown in Figure~\ref{fig:AMG}, we use the first 10 layers of trained VGG-16 model \cite{simonyan2014very} as the front end to extract low-level features. We build the back end by adopting multiple dilated convolution layers of different dilation rates with an architecture similar to the inception module in \cite{szegedy2015going}. The multiple dilated convolution architecture is motivated
from \cite{wei2018revisiting}. It has the capability of localizing people clusters with enlarged receptive fields. The inception module was originally proposed in \cite{szegedy2015going} to process and aggregate visual information of various scales. We use this module to deal with the diversified crowd distribution in congested scenes.

\subsection{Density Map Estimator}
\begin{figure}[]
	\centering
	\includegraphics[width=0.46\textwidth]{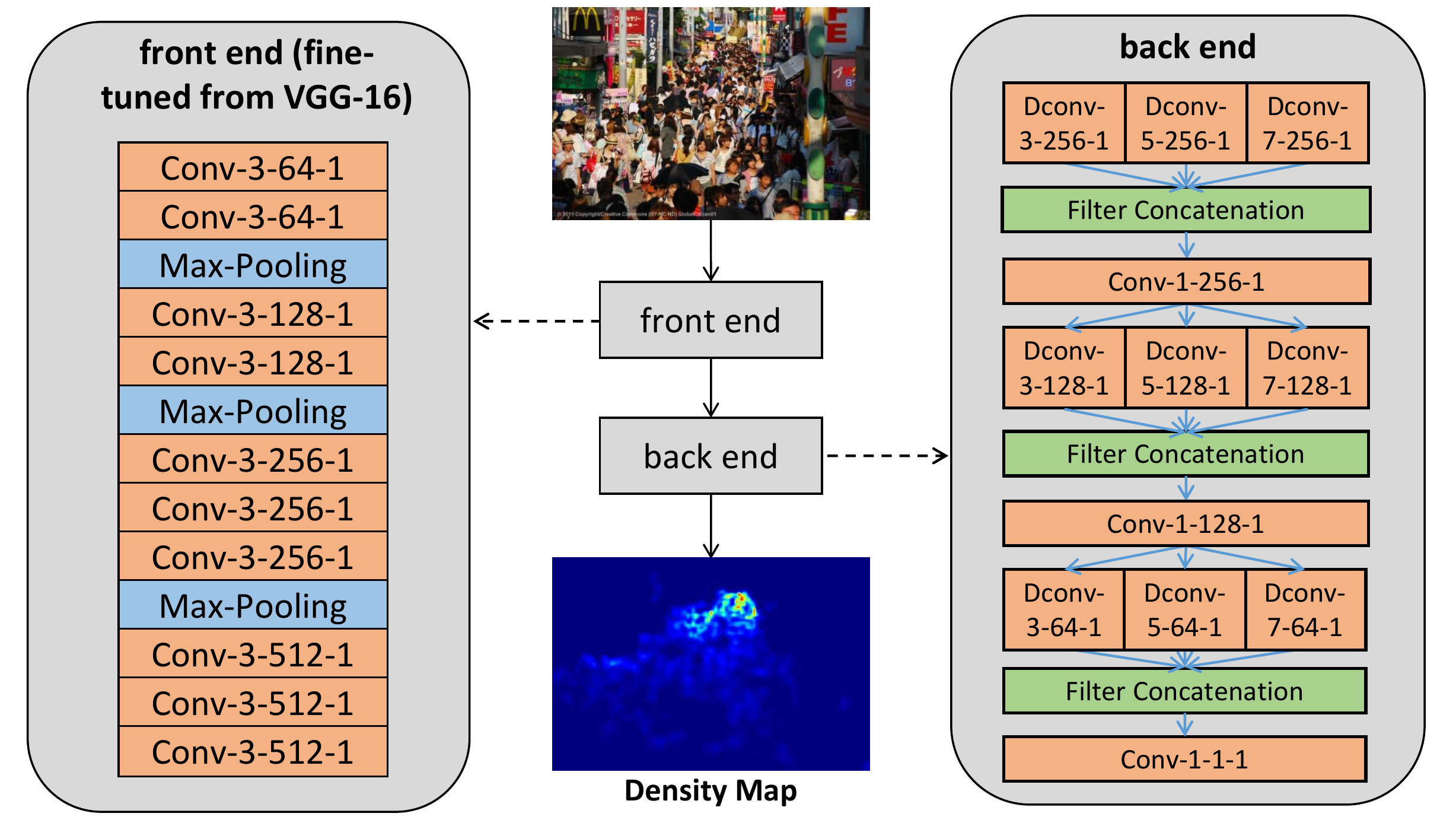}
	\caption{Architecture of DME. The convolutional layers' parameters are denoted as ``Conv-(kernel size)-(number of filters)-(stride)'', max-pooling layers are conducted over a 2$\times$2 pixel window, with stride 2. The deformable convolutional layers' parameters are denoted as {``Dconv-(kernel size)-(number of filters)-(stride)''}.}
	\label{fig:DMEArchi}
\end{figure}

\noindent The DME network consists of two components: the front end and the back end. We remove the fully-connected layers of VGG-16 \cite{simonyan2014very} and leave 10 convolutional layers to as the front end of the DME. The back end is a multi-scale deformable convolution based CNN network \cite{dai2017deformable}. The architecture of DME is shown in Figure~\ref{fig:DMEArchi}. The front end uses the first 10 layers of trained VGG-16 model~\cite{simonyan2014very} to extract low-level features. The back end uses multi-scale deformable convolutional layers with a structure similar to the inception module in \cite{szegedy2015going}, which enables DME to cope with various occlusion, diversified crowd distribution, and the distortion caused by perspective view.

\begin{figure}[ht]
	\centering
	\includegraphics[width=0.4\textwidth]{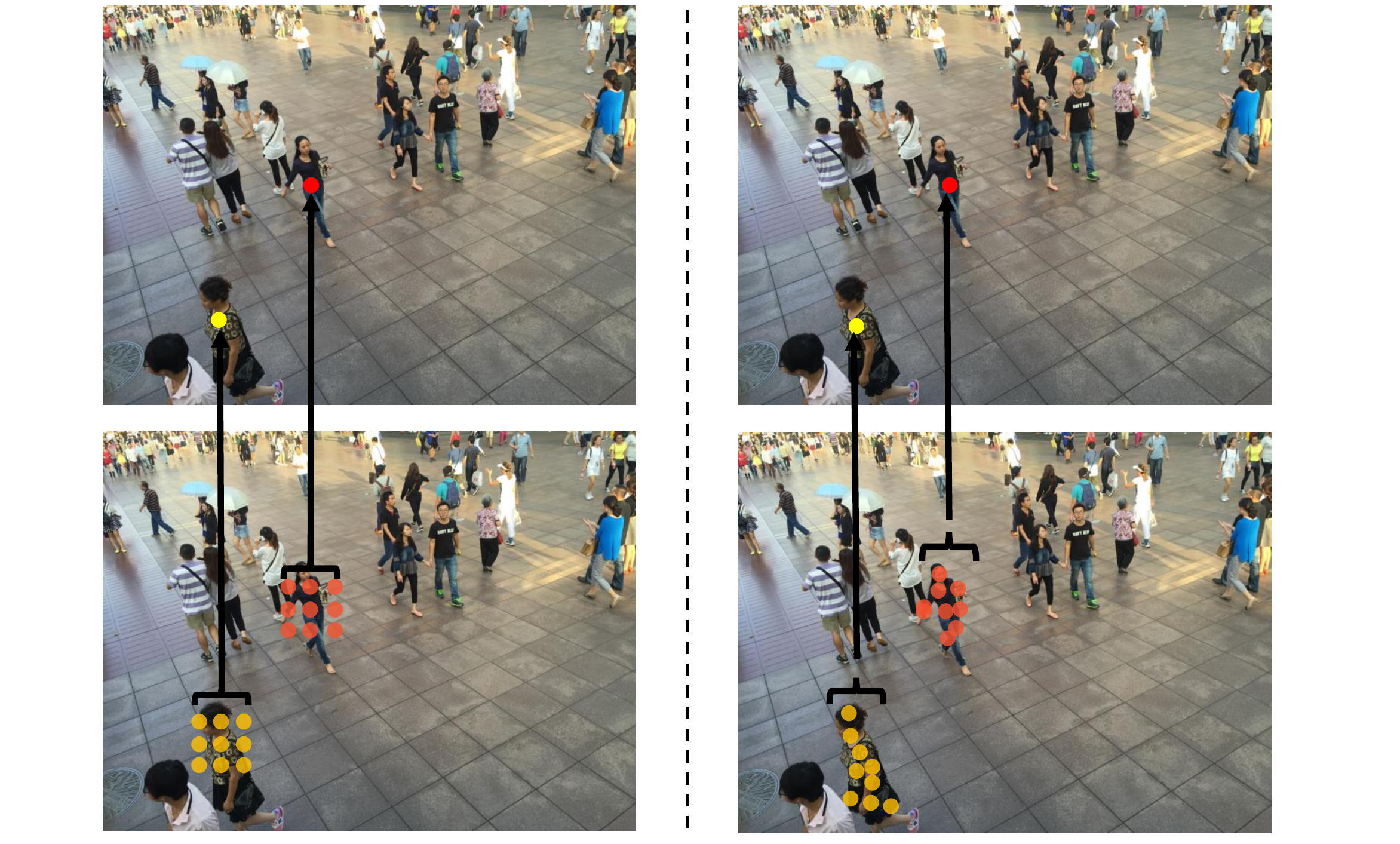}
	\caption{Illustration of the deformed sampling locations. Left: standard convolution; right: deformable convolution; top: activation units on the feature map; bottom: the sampling locations of the $3 \times 3$ filter.}
	\label{fig:dcov_example}
\end{figure}

The deformable convolution scheme was originally proposed in \cite{dai2017deformable}. Beneficial from the adaptive (deformable) sampling location selection scheme,  deformable convolution has shown its effectiveness on various tasks, such as object detection, in the wild environment. The deformable convolution treats the offsets of sampling locations as learning parameters. Rather than
uniform sampling, the sampling locations in the deformable convolution can be adjusted and optimized via training (see Figure~\ref{fig:dcov_example} for the deformed sampling points by the deformable convolution on an example form ShanghaiTech Part\_B dataset ~\cite{zhang2016single}). Compared to the uniform sampling scheme, this kind of dynamic sampling scheme is more suitable for the crowd understanding problem of congested noisy scenes. We will show the comparative advantages of the deformable convolution in our experimental section.

\section{Experiments}\label{sec:exp}
\subsection{Datasets and Settings}
We evaluate \OurNet{} on four challenging datasets for crowd counting: ShanghaiTech dataset~\cite{zhang2016single}, the UCF\_CC\_50 dataset~\cite{idrees2013multi}, the WorldExpo'10 dataset~\cite{zhang2015cross}, and the UCSD dataset~\cite{chan2008privacy}.

\textbf{ShanghaiTech dataset} \cite{zhang2016single}. The ShanghaiTech dataset contains 1,198 images with a total of 330,165 people. It is divided into two parts: Part\_A and Part\_B. Part\_A contains 482 pictures of congested scenes, in which 300 are used as training dataset and 182 are used as testing dataset; Part\_B contains 716 images of sparse scene, 400 of which are used as training dataset and 316 are used as testing dataset.

\textbf{UCF\_CC\_50 dataset} \cite{idrees2013multi}. This dataset contains 50 images downloaded from the Internet. The number of persons per image ranges from 94 to 4543 with an average of 1280 individuals. It is a very challenging dataset with two problems: the limited number of the images and the large span in person count between images. We used 5-fold-cross-validation setting described in \cite{idrees2013multi}.

\textbf{WorldExpo'10 dataset} \cite{zhang2015cross}. It contains 3980  from 5 different scenes. Among 3980 images, 3380 images are used as training dataset and the remaining 600 images are used as testing dataset.
Region-of-Interest (ROI) regions are provided in this dataset.

\textbf{UCSD dataset} \cite{chan2008privacy}. The UCSD dataset contains 2000 images in sparse scene. The dataset also provides ROI region information. We created the ground truth in the same way as we did for the WorldExpo'10 dataset. Since the size of each image is too small to support the generation of high-quality density maps, we therefore enlarge each image to 952$\times$632 size by bilinear interpolation. Among the 2000 images, 800 images were used as training dataset, and the rest were used as testing dataset.
Region-of-Interest (ROI) regions are also provided in this dataset.

\begin{figure*}[]
	\centering
	\includegraphics[width=0.9\textwidth]{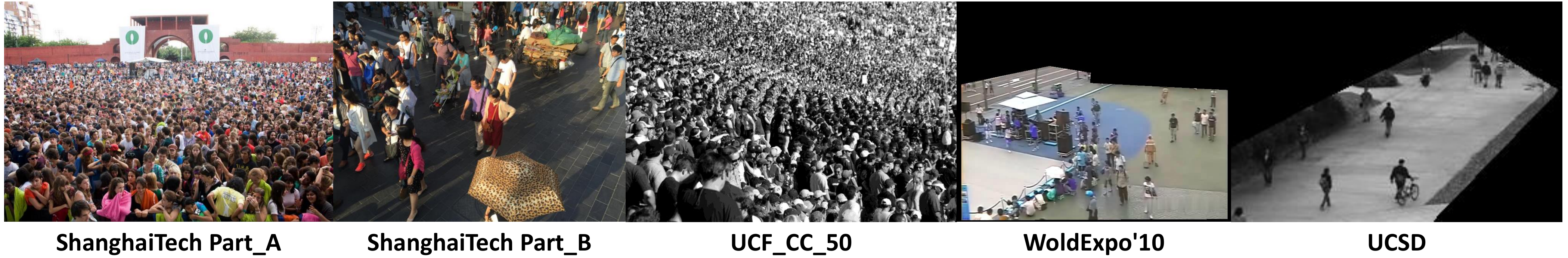}
	\caption{Representative examples from four crowd counting datasets.}
	\label{fig:examples}
	\vspace{-0.1in}
\end{figure*}

We show a representative example for each crowd counting dataset in Figure~\ref{fig:examples}. These four crowd counting datasets have their own characteristics. In general, the scenes in ShanghaiTech Part\_A dataset
are congested and noisy. Examples in ShanghaiTech Part\_B are noisy but not highly congested. The UCF\_CC\_50 dataset consists of extremely congested scenes which have
hardly any background noises. Both WorldExpo'10 dataset and UCSD dataset provide example with sparse crowd scenes in the form of ROI regions.
Scenes in the ROI regions of the WorldExpo'10 dataset are generally noisier than the only one scene in the UCSD dataset.




Following~\cite{sindagi2017generating,li2018csrnet}, we use the mean absolute error (MAE) and the mean square error (MSE) for quantitative evaluation of the estimated density maps.
PSNR (Peak Signal-to-Noise Ratio) and SSIM~\cite{wang2004image} are used to measure the quality of the generated density map.
For fair comparison, we follow the measurement procedure in \cite{li2018csrnet} and resize the density map and ground truth to the size of the original input image by linear interpolation.

\subsection{Training}
\subsubsection{AMG Training}
Training data for the binary classification network AMG consists of two groups of samples: positive and negative samples. The positive samples are from the training sets of the four crowd counting datasets. The negative samples are 650 background images downloaded from the Internet. These negative samples are shared by the training of each individual dataset. These 650 negative samples contain various outdoor scenes where people appear, such as streets, squares, etc., ensuring that the biggest difference between positive sample and negative samples is whether the image contains people. Adam~\cite{kingma2014adam} is selected as the optimization method with the learning rate at 1e-5 and Standard cross-entropy loss is used as the loss function.

\subsubsection{DME training}
We simply crop 9 patches from each image where each patch is 1/4 of the original image size. The first four patches contain four quarters of the image without overlapping. The other five patches are randomly cropped from the image. After that, we mirror the patches so that we double the training dataset. We generate the ground truth for DME training following the procedure in~\cite{li2018csrnet}. We select Adam \cite{kingma2014adam} as the optimization method with the learning rate at 1e-5. As previous works \cite{zhang2016single,sam2017switching,li2018csrnet}, we use the euclidean distance to measure the difference between the generating density map and ground truth and define the loss function as

$$L(\Theta)=\frac{1}{2N}\sum_{i=1}^N ||F(X_i;\Theta)-F_i||_{2}^2\eqno(2),$$
where $N$ is the batch size, $F(X_i;\Theta)$ is the estimated density map generated by DME with the parameter$\Theta$, $X_i$ is the input image, and $F_i$ is the ground truth of $X_i$.

\subsection{Results and Analyses}	
In this section, we first study several alternative network design of \OurNet{}. After that, we evaluate the overall performance of \OurNet{} and compare it with previous state-of-the-art methods.  

\subsubsection{Alternative study}
\begin{table}[!ht]
	\centering
	\resizebox{82mm}{10.5mm}{
		\begin{tabular}{|c|c|c|c|c|c|c|c|c|}
			\hline
			\multicolumn{1}{|c|}{ } & \multicolumn{2}{c|}{\textbf{DME}} & \multicolumn{2}{c|}{\textbf{AMG-DME}} &  \multicolumn{2}{c|}{\textbf{AMG-bAttn-DME}} &
			\multicolumn{2}{c|}{\textbf{AMG-attn-DME}}\\
			\hline\hline
			Dataset&MAE&MSE&MAE&MSE&MAE&MSE&MAE&MSE\\
			\hline
			ShanghaiTech Part\_A~\cite{zhang2016single}&68.5&107.5&66.1&102.1&\textbf{63.2}&\textbf{98.9}&70.9&115.2\\
			\hline
			ShanghaiTech Part\_B~\cite{zhang2016single}&9.3&16.9&\textbf{7.6}&13.9&8.2&15.7&7.7&\textbf{12.9}\\
			\hline
			UCF\_CC\_50~\cite{idrees2013multi}&\textbf{257.1}&363.5&257.9&\textbf{357.7}&266.4&358.0&273.6&362.0\\
			\hline
			The WorldExpo'10~\cite{zhang2015cross}&8.5&-&7.4&-&7.7&-&\textbf{7.3}&-\\
			\hline
			The UCSD~\cite{chan2008privacy}&\textbf{0.98}&\textbf{1.25}&1.10&1.42&1.39&1.68&1.09&1.35\\
			\hline
	\end{tabular}}
	\caption{Results of different variants of \OurNet{} on four crowd counting datasets.}\label{tab:ablation}
\end{table}

\begin{figure}[!ht]
	\centering
	\includegraphics[ width=0.48\textwidth]{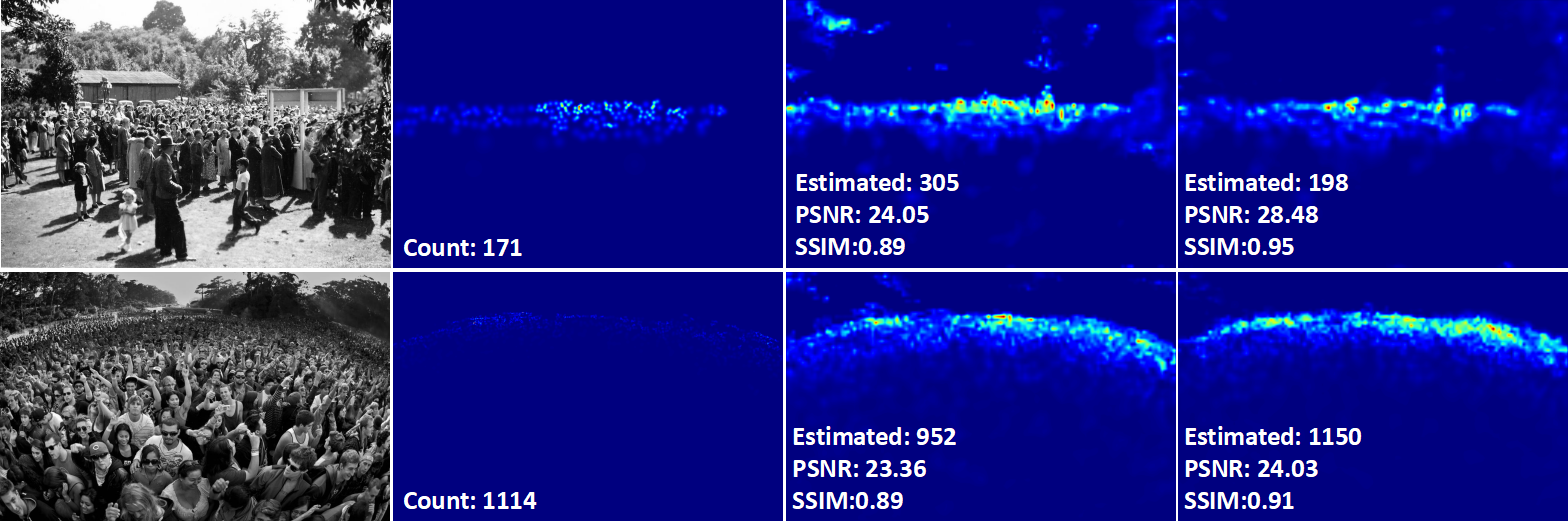}
	\caption{DME vs. AMG-DME. From left to right: representative samples from the ShanghaiTech Part\_A dataset, ground truth density map, density map generated from the architectures of single DME and AMG-DME.}
	\vspace{-0.1in}
	\label{fig:DMEVSAMG}
\end{figure}
\textbf{DME}. Our first study is to investigate the influence of the AMG network, we compared two network designs on all the four datasets. The first one named AMG-DME has the architecture shown in Figure~\ref{fig:overviewArch}. The other one named DME uses the only DME network. Our quantitative experimental results in Table~\ref{tab:ablation} show that AMG-DME is significantly superior than DME on the those datasets which are characteristic of noisy scenes: ShanghaiTech Part\_A, Part\_B and WorldExpo'10. In Figure~\ref{fig:DMEVSAMG}, we illustrate two representative samples from the testing set of ShanghaiTech Part\_A. On the top example which contains a congested noisy scene, estimated people number of AMG-DME is 198 that is much closer to the groud truth 171 than
that estimated by DME. From the density map in the 3rd column of
Figure~\ref{fig:DMEVSAMG}, we can see the trees in the distance have been recognized as people by the single DME model. However, AMG-DME does not suffer this problem due to the help from the AMG network. On the middle-row example containing a noisy and more congested scene, the performances of AMG-DME and DME agree with those on the top example.  The comparison results indicate that AMG-DME is more effective than DME on those noisy examples. 

On the UCF\_CC\_50 dataset, AMG-DME has approximate performance (slightly higher AME but lower MSE) with DME. It may due to
the fact most of examples in the UCF\_CC\_50 dataset have a large regions of congested crowds while rarely have background noises.
On the UCSD dataset where scenes are neither congested nor noisy, both MSE and MAE of AMG-DME is slightly higher than DME. This
might because the examples in the UCSD dataset have already provide the accurate information of ROI regions. The attention map generated by the AMG network
may destroy the ROI regions, which degrades the performance of the DME network since some ROI regions may be erased from its input.

\begin{figure}[ht]
	\centering
	\includegraphics[ width=0.45\textwidth]{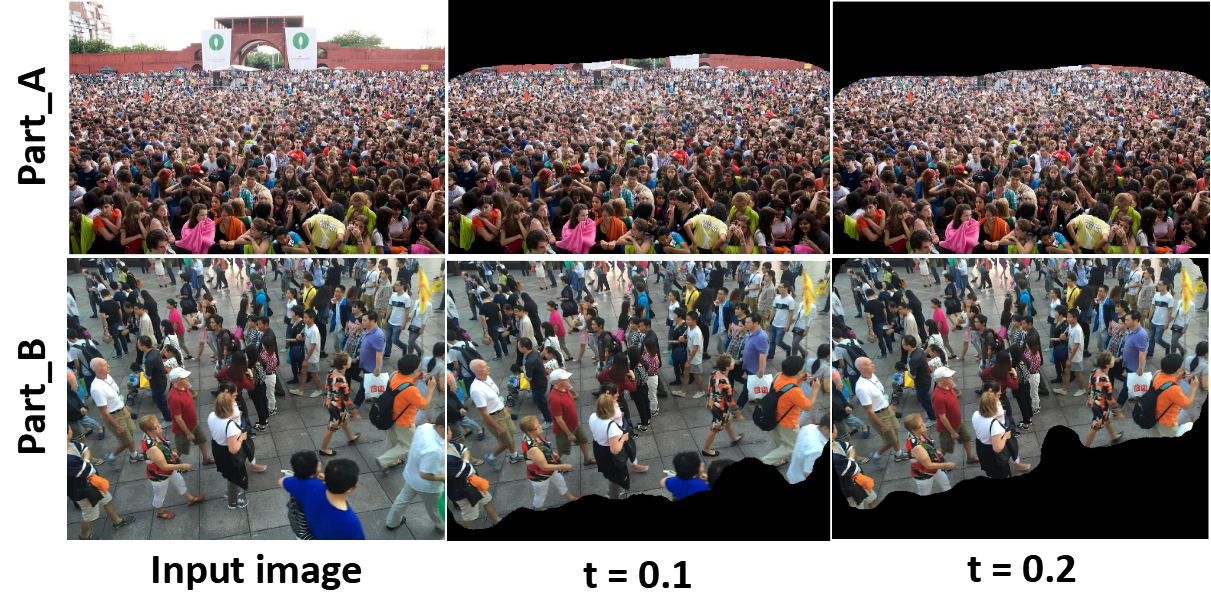}
	\caption{Illustration of the ROI regions extracted by different attention thresholds. The pixels which have the value of attention lower than $t$ are changed to black.}
	\label{fig:partA_seg}
\end{figure}

\textbf{AMG-bAttn-DME}. 
Since the AMG network has shown its strength in coping with noise background of scenes, our second study is to explore
if a hard binary attention mask is more effective than the soft attention employed by AMG-DME. We therefore set up an variant of AMG-DME called AMG-bAttn-DME in Table~\ref{tab:ablation}.
AMG-bAttn-DME has the same architecture as AMG-DME while differing with AMG-DME on the attention map (i.e., the attention maps of AMG-bAttn-DME contain either 0 or 1, other
than a floating point within $[0,1]$ in the attention maps of AMG-DME).
We first conducted the experiments on the ShanghaiTech dataset to find out the optimal binarization threshold for AMG-bAttn-DME.
We set three different threshold attention values,$\{0.2, 0.1, 0.0\}$, for the binarization of attention maps.
The ROI regions are gradually enlarged with the decreasing the threshold values as shown in Figure~\ref{fig:partA_seg}.
The results shown in Table~\ref{tab:sh_thre} indicates AMG-bAttn-DME with attention threshold of 0.1 achieved the best performance.
We then evaluated AMG-bAttn-DME with this optimal attention threshold on the rest three datasets and reported the results in Table~\ref{tab:ablation}.
It is observed that AMG-bAttn-DME is superior than AMG-DME only on ShanghaiTech Part\_A while AMG-DME outperforms AMG-bAttn-DME on all other datasets.
It may be due to the AMG network can learn more accurate attention maps on ShanghaiTech Part\_A and the binarization process does not
destroy too much information of the crowd regions.

\begin{table}[!ht]
	\centering
	\small
	\begin{tabular}{|c|c|c|c|c|}
		\hline
		\multicolumn{1}{|c|}{ } & \multicolumn{2}{c|}{\textbf{Part\_A}}&
		\multicolumn{2}{c|}{\textbf{Part\_B}}\\
		\hline\hline
		Threshold&MAE&MSE&MAE&MSE\\
		\hline
		t = 0.2&68.0&104.1&9.2&17.8\\
		\hline
		t = 0.1&63.2&98.9&8.2&15.7\\
		\hline
		t = 0.0 &63.2&100.6&8.6&15.0\\
		\hline
	\end{tabular}
	\caption{Performance of AMG-bAttn-DME under different binarization thresholds on the ShanghaiTech dataset.}\label{tab:sh_thre}
\end{table}

\begin{figure}[ht!]
	\centering
	\includegraphics[width=0.45\textwidth]{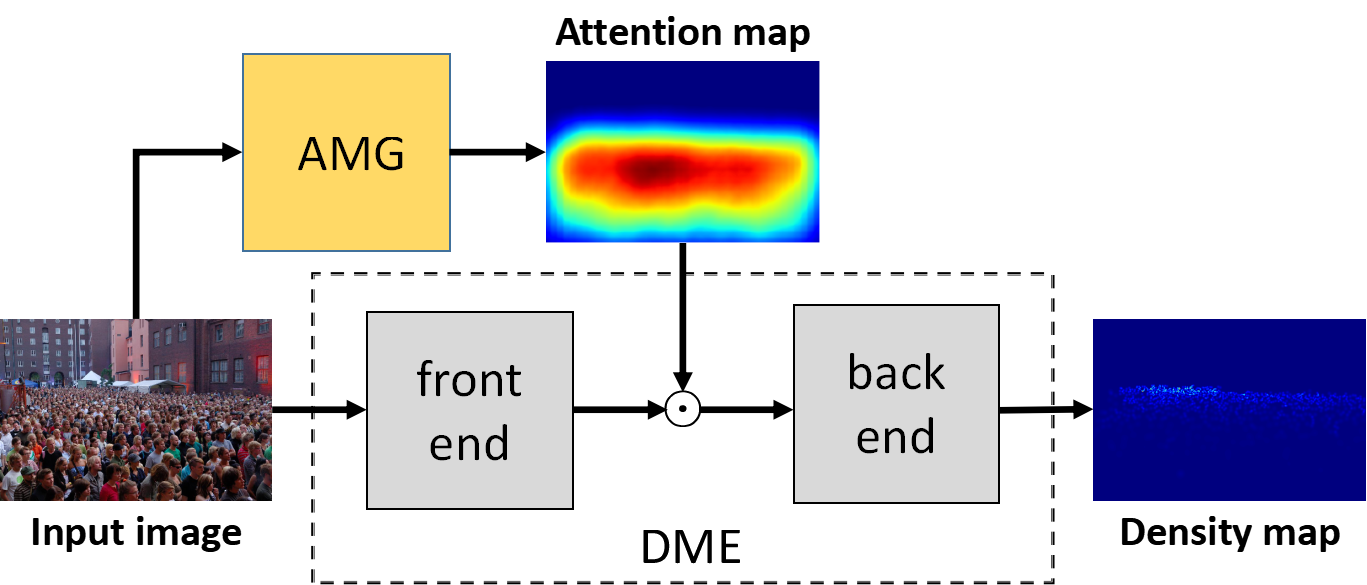}
	\caption{Architecture of AMG-attn-DME in both training and testing phases.}
	\vspace{-0.15in}
	\label{fig:model_AMG-attn-DME}
\end{figure}

\textbf{AMG-attn-DME}.
Complement to the above experiments, we stretched the design choice exploration to studying an alternative way of
injecting the learned attention from the AMG network to the DME network. In our proposed architecture, the DME
network directly takes the crowd images as input. An alternative architecture is to weigh intermediate
the feature map of a certain layer of the DME network with the attention map from the AMG network.
In our implementation, we inject the attention map into the output of the front end of the DME network as shown in Figure~\ref{fig:model_AMG-attn-DME}. Following the same training procedures as those in Table~\ref{tab:ablation}, this alternative architecture,
named AMG-attn-DME, performs slightly worse than AMG-DME on the datasets with congested noisy scenes like
ShanghaiTech Part\_A and ShanghaiTech Part\_B. This may be due to some non-crowd pixels in the attention map from
the AMG network having an attention value of zero, which, during the injection, would make
convolution features at those corresponding locations vanish, reducing the feature information learned
by previous convolutional lays from the input. On the UCF\_CC\_50 dataset and UCSD dataset, AMG-attn-DME is worse than the the only DME network as
AMG-bAttn-DME and AMG-DME. This is because the scenes of these two datasets have less noisy background, AMG-attn-DME may reduce the information of the ROI regions through the injected attention map.
On the UCSD and WorldExpo'10 datasets, AMG-attn-DME achieved higher effectiveness. Maybe it is because the convolution feature vanishing problem
has been alleviated by the black regions around the ROI regions in the input.

\subsubsection{Quantitative results}
In this section, we study the overall performance of \OurNet{} and compare it with existing methods on each individual crowd counting dataset.

\textbf{Comparison on MAE and MSE}.
We first compare the variants of the proposed \OurNet{} network with the state-of-the-art work CSRNet~\cite{li2018csrnet} along with
several previous methods including CP-CNN~\cite{sindagi2017generating}, MCNN~\cite{zhang2016single}, Cascaded-MTL~\cite{sindagi2017cnn},
Switching-CNN~\cite{sam2017switching} on the ShanghaiTech dataset and the UCF\_CC\_50 dataset.
These two datasets are characteristic of congested and/or noisy scenes. The comparison results were summarized in Table~\ref{tab:SHTPA}. On the ShanghaiTech dataset, two of our approach variants \OurNet{}(AMG-DME) and \OurNet{}(AMG-bAttn-DME)
achieved better performances than existing approaches. The only DME network achieved the performance generally close to the state-of-the-art approach CSRNet~\cite{li2018csrnet}.

%

\begin{table}[!ht]
	\centering
	\resizebox{82mm}{17.5mm}{
		\begin{tabular}{|c|c|c|c|c|c|c|}
			\hline
			\multicolumn{1}{|c|}{ } & \multicolumn{2}{c|}{\textbf{Part\_A}} & \multicolumn{2}{c|}{\textbf{Part\_B}}&
			\multicolumn{2}{c|}{\textbf{UCF\_CC\_50}}\\
			\hline\hline
			Method & MAE & MSE & MAE & MSE& MAE & MSE\\
			\hline
			MCNN~\cite{zhang2016single} & 110.2 & 173.2 & 26.4 & 41.3& 377.6 & 509.1\\
			\hline
			Cascaded-MTL~\cite{sindagi2017cnn} & 101.3 & 152.4 & 20.0 & 31.1&322.8&397.9\\
			\hline
			Switching-CNN~\cite{sam2017switching} & 90.4 & 135.0 & 21.6 & 33.4&318.1&439.2\\
			\hline
			CP-CNN~\cite{sindagi2017generating} & 73.6 & 106.4 & 20.1 & 30.1&295.8&\textbf{320.9}\\
			\hline
			CSRNet~\cite{li2018csrnet} & 68.2 & 115.0 & 10.6 & 16.0&266.1&397.5\\
			\hline
			\OurNet{}(DME) &68.5&107.5&9.3&16.9&\textbf{257.1}&363.5\\
			\hline
			\OurNet{}(AMG-DME) &66.1&102.1&\textbf{7.6}&13.9&257.9&357.7\\
			\hline
			\OurNet{}(AMG-bAttn-DME) & \textbf{63.2} & \textbf{98.9} & 8.2 & 15.7&266.4&358.0\\
			\hline
			\OurNet{}(AMG-attn-DME) &70.9&115.2&7.7&\textbf{12.9}&273.6&362.0\\
			\hline
	\end{tabular}}
	\caption{Estimation errors on ShanghaiTech and UCF\_CC\_50.}\label{tab:SHTPA}
\end{table}

\begin{table}[!ht]
	\centering
	\resizebox{82mm}{12.5mm}{
		\begin{tabular}{|c|c|c|c|c|c|c|c|c|}
			\hline
			\multicolumn{1}{|c|}{ }&
			\multicolumn{6}{c|}{\textbf{The WorldEXpo'10}} & \multicolumn{2}{c|}{\textbf{UCSD}}\\
			\hline\hline
			Method&Sce.1&Sce.2&Sce.3&Sce.4&Sce.5&Ave.&MAE&MSE\\
			\hline
			MCNN~\cite{zhang2016single}&3.4&20.6&12.9&13.0&8.1&11.6&1.07&1.35\\
			\hline
			Switching-CNN~\cite{sam2017switching}&4.4&15.7&10.0&11.0&5.9&9.4&1.62&2.10\\
			\hline
			CSRNet~\cite{li2018csrnet}&2.9&\textbf{11.5}&\textbf{8.6}&16.6&3.4&8.6&1.16&1.47\\
			\hline
			\OurNet{}(DME)&\textbf{1.6}&15.8&11.0&10.9&3.2&8.5&\textbf{0.98}&\textbf{1.25}\\
			\hline
			\OurNet{}(AMG-DME)&\textbf{1.6}&13.8&10.7&8.0&3.2&7.4&1.10&1.42\\
			\hline
			\OurNet{}(AMG-bAttn-DME)&1.7&14.4&11.5&\textbf{7.9}&3.0&7.7&1.39&1.68\\
			\hline
			\OurNet{}(AMG-attn-DME)&\textbf{1.6}&13.2&8.7&10.6&\textbf{2.6}&\textbf{7.3}&1.09&1.35\\
			\hline
	\end{tabular}}
	\caption{Estimation error comparison on the WorldExpo'10 and UCSD. Note that only MAE is provided on WorldExpo'10 as previous approaches.}\label{tab:UCSD}
\end{table}
On the two relatively less challenging datasets WorldExpo'10 and UCSD, we compared \OurNet{} with
recent state-of-the-art recent approaches including Switching-CNN~\cite{sam2017switching}, MCNN~\cite{zhang2016single}, and CSRNet~\cite{li2018csrnet}.
The comparison results are shown in Table~\ref{tab:UCSD}.
Our method achieved the best accuracy in scenes 1, 4, 5 as well as the best average accuracy on the WorldExpo'10 dataset .
On the UCSD dataset, our DME model achieved the best accuracy on terms of both MAE and MSE.

\textbf{Comparison on PSNR and SSIM.}
To study the quality of the density maps generated by \OurNet{}, another experiment was conducted on all the five datasets for both \OurNet{} and the state-of-the-art method CSRNet~\cite{li2018csrnet}. The comparison results are shown in Table \ref{tab:qulaityCSR}. Our method outperforms CSRNet on all the five datasets. On UCF\_CC\_50 dataset, our method improves 7.03\% on PSNR and 55.76\% on SSIM. On USCD dataset, our method improves 31.81\% on PSNR and 8.13\% on SSIM.
\begin{table}[!ht]
	\centering
	\resizebox{82mm}{17mm}{
		\begin{tabular}{|c|c|c|c|c|}
			\hline
			\multicolumn{1}{|c|}{ } & \multicolumn{2}{c|}{\textbf{CSRNet~\cite{li2018csrnet}}} & \multicolumn{2}{c|}{\textbf{\OurNet{}}}\\
			\hline\hline
			Dataset&PSNR&SSIM&PSNR&SSIM\\
			\hline
			ShanghaiTech Part\_A~\cite{zhang2016single}&23.79&0.76&\textbf{24.48}&\textbf{0.88}\\
			\hline
			ShanghaiTech Part\_B~\cite{zhang2016single}&27.02&0.89&\textbf{29.35}&\textbf{0.97}\\
			\hline
			UCF\_CC\_50~\cite{idrees2013multi}&18.76&0.52&\textbf{20.08}&\textbf{0.81}\\
			\hline
			The WorldExpo'10~\cite{zhang2015cross}&26.94&0.92&\textbf{29.12}&\textbf{0.95}\\
			\hline
			The UCSD~\cite{chan2008privacy}&20.02&0.86&\textbf{26.39}&\textbf{0.93}\\
			\hline
			TRANCOS~\cite{guerrero2015extremely}&27.10&0.93&\textbf{29.56}&\textbf{0.97}\\
			\hline
	\end{tabular}}
	\caption{ CSRNet vs. \OurNet{} (AMG-DME).}\label{tab:qulaityCSR}
\end{table}

\textbf{{Evaluation on vehicle counting dataset.}}
We conducted experiments on the TRANCOS~\cite{guerrero2015extremely}  dataset for vehicle counting to evaluate the generalization capability of the proposed approach.
The positive samples for training are from the training set of TRANCOS~\cite{guerrero2015extremely}. The negative samples use 250 background images downloaded from the Internet, including various road scenes without vehicle. 
As previous work CSRNet~\cite{li2018csrnet}, we use the Grid Average Mean Absolute Error (GAME) to measure the counting accuracy. The comparison results are shown in Table \ref{tab:TRANCOS}. It clearly shows that the \OurNet{} approach achieved the best performance at all levels of GAMEs.  



\begin{table}[!ht]
	\centering
	\resizebox{82mm}{13.5mm}{
		\begin{tabular}{|c|c|c|c|c|}
			\hline
			\textbf{Method}&\textbf{GAME0}&\textbf{GAME1}&\textbf{GAME2}&\textbf{GAME3}\\
			\hline\hline
			Hydra-3s\cite{onoro2016towards}&10.99&13.75&16.69&19.32\\
			\hline
			FCN-HA~\cite{zhang2017fcn}&4.21&-&-&-\\
			\hline
			CSRNet~\cite{li2018csrnet}&3.56&5.49&8.75&15.04\\
			\hline
			\OurNet{}(DME)&2.65&4.49&7.09&14.29\\
			\hline
			\OurNet{}(AMG-DME)&\textbf{2.39}&4.23&6.89&14.82\\
			\hline
			\OurNet{}(AMG-bAttn-DME) &2.69&4.61&7.13&14.14\\
			\hline
			\OurNet{}(AMG-attn-DME)&2.44&\textbf{4.14}&\textbf{6.78}&\textbf{13.58}\\
			\hline
	\end{tabular}}
	\caption{Evaluation on TRANCOS.}\label{tab:TRANCOS}
\end{table}

\vspace{-0.1in}
\subsubsection{Qualitative results}
In this section, we further investigate the general performance  of the proposed \OurNet{} by qualitative results. We mainly compared \OurNet{}  with the state-of-the-art approach CSRNet~\cite{li2018csrnet} which have demonstrated the best performance on the datasets including the ShanghaiTech, UCF\_CC\_50,  the WorlExpo'10, and UCSD datasets. In general, CSRNet has a front-end and back-end  architecture as the DME network of the proposed \OurNet{}.  It is empowered by a dilated convolution design in the back-end of its architecture. Apart from the additional AMG netwok, \OurNet{} differs from CSRNet by two additional features in its DME network: 1) the multiple-scale convolution scheme different from the single scale scheme of CSRNet, and 2) the deformable sampling scheme different from the evenly fixed-offset sampling in the dilated convolution of CSRNet.  

\begin{figure}[ht!]
	\centering
	\includegraphics[width=0.49\textwidth]{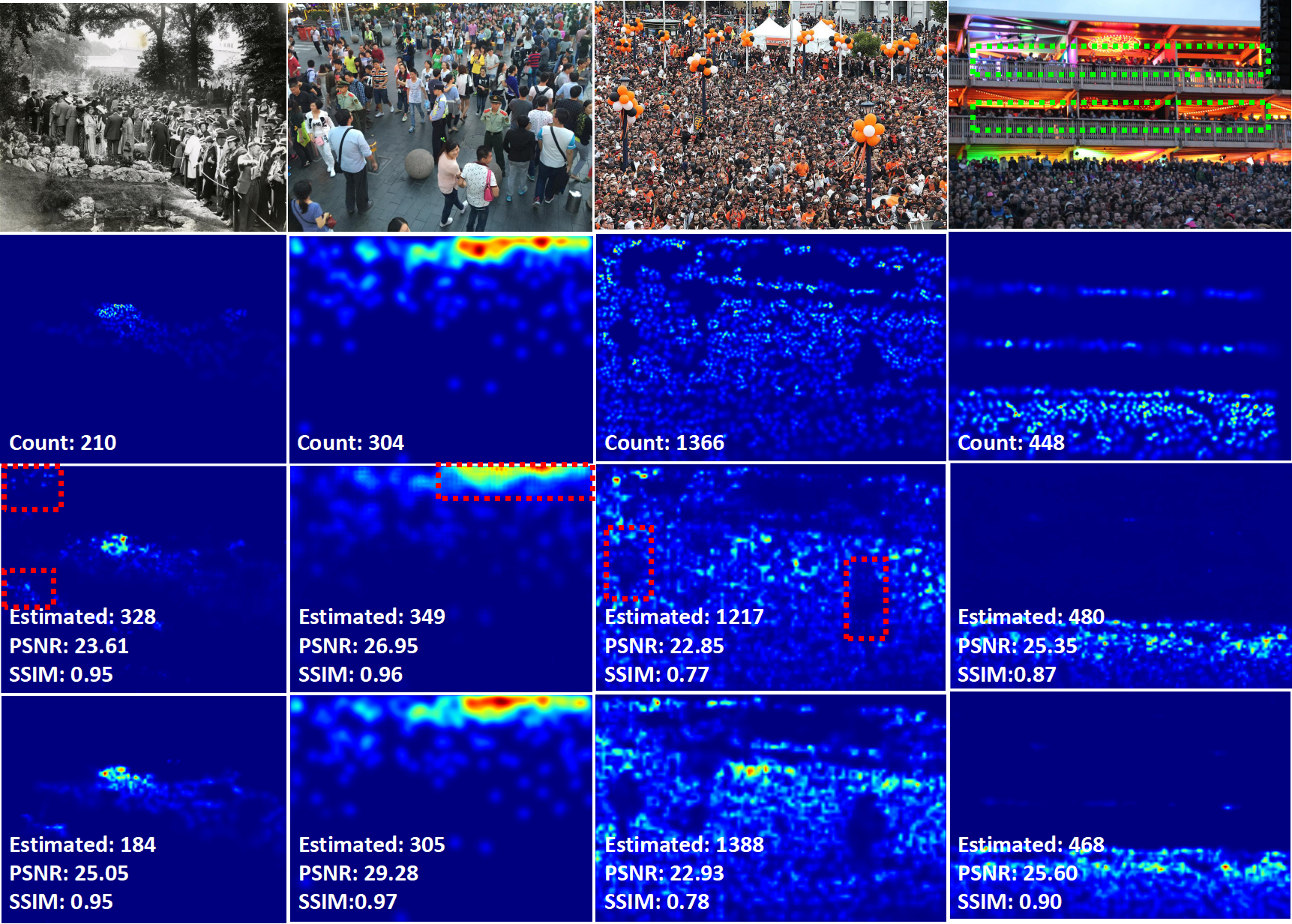}
	\caption{
		From top to bottom: representative samples from the testing set of the ShanghaiTech dataset, ground truth density maps, estimated density maps generated by the state-of-the-art approach CSRNet~\cite{li2018csrnet} and \OurNet{} (AMG-DME)
		respectively. 
	}
	\label{fig:Qualitative}
\end{figure}

Figure~\ref{fig:Qualitative} shows some qualitative comparisons between the proposed \OurNet{} (the variant AMG-DME is used) and the state-of-the-art approach CSRNet~\cite{li2018csrnet}. 
Through visualization, it is observed that CSRNet is much less effective on those examples with various noises than \OurNet{}. We can see the evidence from the noise regions marked by red boxes of the 1st column where noises exist in the background regions, as well as the marked regions of the 3rd column where noises can be found in the crowd regions.  This may be due to CSRNet directly takes the crowd image as input while the DME network of \OurNet{} takes as the input the crowd information highlighted by its AMG network. On the example of the 2nd column where there is not much noise but a significantly non-uniform crowd distribution,  \OurNet{} also clearly outperforms CSRNet.  This indicates that the multi-scale deformable convolution scheme in \OurNet{} is more effective than the single-scale fixed-offset dilated convolution scheme in CSRNet. 

On the rightmost example of Figure~\ref{fig:Qualitative}  which 
have highly occluded crowd regions (see the regions within the two green dotted bordered rectangle), \OurNet{} only recognized part of the severely occluded crowd regions. It may because the AMG network of \OurNet{} cannot highlight out the whole occluded crowd regions for the DME network.  Nevertheless, \OurNet{} still achieved better performance in terms of all the measurement parameters: estimated number, PSNR and SSIM.

%

%

\section{Conclusion}\label{sec:conclusion}
We propose a convolutional neural network based architecture named \OurNet{} for crowd understanding of congested noisy scenes. Benefiting from the  multi-scale deformable convolutional layers and \cq{attention-aware} training scheme, \OurNet{} generally achieved more accurate crowd counting and density map estimation than existing methods by suppressing the problems caused by noises, occlusions, and diversified crowd distributions commonly presented in highly congested noisy environments. On four popular crowd counting datasets (ShanghaiTech, UCF\_CC\_50, WorldEXPO'10, UCSD) and an extra vehicle counting dataset TRANCOS, \OurNet{} achieved significant improvements over recent state-of-the-art approaches. 


	{\small
		\bibliographystyle{ieee_fullname}
		\bibliography{mybib}

\begin{thebibliography}{10}\itemsep=-1pt

\bibitem{boominathan2016crowdnet}
Lokesh Boominathan, Srinivas~S.S. Kruthiventi, and R.~Venkatesh Babu.
\newblock Crowdnet: A deep convolutional network for dense crowd counting.
\newblock In {\em Proc. ACM MM}, pages 640--644, 2016.

\bibitem{CaoWZS18eccv}
Xinkun Cao, Zhipeng Wang, Yanyun Zhao, and Fei Su.
\newblock Scale aggregation network for accurate and efficient crowd counting.
\newblock In {\em Proc. Springer ECCV}, pages 757--773, 2018.

\bibitem{chan2008privacy}
Antoni~B Chan, Zhang-Sheng~John Liang, and Nuno Vasconcelos.
\newblock Privacy preserving crowd monitoring: Counting people without people
  models or tracking.
\newblock In {\em Proc. IEEE CVPR}, pages 1--7, 2008.

\bibitem{ChenGXL13}
Ke Chen, Shaogang Gong, Tao Xiang, and Chen~Change Loy.
\newblock Cumulative attribute space for age and crowd density estimation.
\newblock In {\em Proc. IEEE CVPR}, pages 2467--2474, 2013.

\bibitem{ChenLGX12}
Ke Chen, Chen~Change Loy, Shaogang Gong, and Tony Xiang.
\newblock Feature mining for localised crowd counting.
\newblock In {\em Proc. BMVC}, pages 1--11, 2012.

\bibitem{chu2017multi}
Xiao Chu, Wei Yang, Wanli Ouyang, Cheng Ma, Alan~L Yuille, and Xiaogang Wang.
\newblock Multi-context attention for human pose estimation.
\newblock In {\em Proc. IEEE CVPR}, pages 1831--1840, 2018.

\bibitem{dai2017deformable}
Jifeng Dai, Haozhi Qi, Yuwen Xiong, Yi Li, Guodong Zhang, Han Hu, and Yichen
  Wei.
\newblock Deformable convolutional networks.
\newblock In {\em Proc. IEEE ICCV}, pages 764--773, 2017.

\bibitem{dalal2005histograms}
Navneet Dalal and Bill Triggs.
\newblock Histograms of oriented gradients for human detection.
\newblock In {\em Proc. IEEE CVPR}, pages 886--893, 2005.

\bibitem{DollarWSP12}
Piotr Doll{\'{a}}r, Christian Wojek, Bernt Schiele, and Pietro Perona.
\newblock Pedestrian detection: An evaluation of the state of the art.
\newblock {\em {IEEE} Transactions on Pattern Analysis and Machine
  Intelligence}, 34(4):743--761, 2012.

\bibitem{guerrero2015extremely}
Ricardo Guerrero-G{\'o}mez-Olmedo, Beatriz Torre-Jim{\'e}nez, Roberto
  L{\'o}pez-Sastre, Saturnino Maldonado-Basc{\'o}n, and Daniel Onoro-Rubio.
\newblock Extremely overlapping vehicle counting.
\newblock In {\em Proc. Springer IbPRIA}, pages 423--431, 2015.

\bibitem{hu2017squeeze}
Jie Hu, Li Shen, and Gang Sun.
\newblock Squeeze-and-excitation networks.
\newblock In {\em Proc. IEEE CVPR}, pages 7132--7141, 2018.

\bibitem{idrees2013multi}
Haroon Idrees, Imran Saleemi, Cody Seibert, and Mubarak Shah.
\newblock Multi-source multi-scale counting in extremely dense crowd images.
\newblock In {\em Proc. IEEE CVPR}, pages 2547--2554, 2013.

\bibitem{KangC18bmvc}
Di Kang and Antoni~B. Chan.
\newblock Crowd counting by adaptively fusing predictions from an image
  pyramid.
\newblock In {\em Proc. BMVC}, 2018.

\bibitem{kingma2014adam}
Diederik~P Kingma and Jimmy Ba.
\newblock Adam: A method for stochastic optimization.
\newblock In {\em Proc. ICLR}, 2015.

\bibitem{lempitsky2010learning}
Victor Lempitsky and Andrew Zisserman.
\newblock Learning to count objects in images.
\newblock In {\em Proc. NIPS}, pages 1324--1332, 2010.

\bibitem{LiHH2018arXiv}
Hanhui Li, Xiangjian He, Hefeng Wu, Saeed~Amirgholipour Kasmani, Ruomei Wang,
  Xiaonan Luo, and Liang Lin.
\newblock Structured inhomogeneous density map learning for crowd counting.
\newblock {\em arXiv preprint arXiv:1801.06642}, 2018.

\bibitem{li2018csrnet}
Yuhong Li, Xiaofan Zhang, and Deming Chen.
\newblock {CSRNet}: Dilated convolutional neural networks for understanding the
  highly congested scenes.
\newblock In {\em Proc. IEEE CVPR}, pages 1091--1100, 2018.

\bibitem{liu2018crowd}
Lingbo Liu, Hongjun Wang, Guanbin Li, Wanli Ouyang, and Liang Lin.
\newblock Crowd counting using deep recurrent spatial-aware network.
\newblock In {\em Proc. IJCAI}, pages 849--855, 2018.

\bibitem{marsden2016fully}
Mark Marsden, Kevin McGuinness, Suzanne Little, and Noel~E O'Connor.
\newblock Fully convolutional crowd counting on highly congested scenes.
\newblock {\em arXiv preprint arXiv:1612.00220}, 2016.

\bibitem{onoro2016towards}
Daniel Onoro-Rubio and Roberto~J L{\'o}pez-Sastre.
\newblock Towards perspective-free object counting with deep learning.
\newblock In {\em Proc. Springer ECCV}, pages 615--629, 2016.

\bibitem{pham2015count}
Viet-Quoc Pham, Tatsuo Kozakaya, Osamu Yamaguchi, and Ryuzo Okada.
\newblock Count forest: Co-voting uncertain number of targets using random
  forest for crowd density estimation.
\newblock In {\em Proc. IEEE ICCV}, pages 3253--3261, 2015.

\bibitem{qian2018attentive}
Rui Qian, Robby~T Tan, Wenhan Yang, Jiajun Su, and Jiaying Liu.
\newblock Attentive generative adversarial network for raindrop removal from a
  single image.
\newblock In {\em Proc. IEEE CVPR}, pages 2482--2491, 2018.

\bibitem{qiu2019crowd}
Zhilin Qiu, Lingbo Liu, Guanbin Li, Qing Wang, Nong Xiao, and Liang Lin.
\newblock Crowd counting via multi-view scale aggregation networks.
\newblock In {\em Proc. IEEE ICME}, 2019.

\bibitem{ren2017end}
Mengye Ren and Richard~S Zemel.
\newblock End-to-end instance segmentation with recurrent attention.
\newblock In {\em Proc. IEEE CVPR}, pages 21--26, 2017.

\bibitem{sam2017switching}
Deepak~Babu Sam, Shiv Surya, and R~Venkatesh Babu.
\newblock Switching convolutional neural network for crowd counting.
\newblock In {\em Proc. IEEE CVPR}, pages 4031--4039, 2017.

\bibitem{shang2016end}
Chong Shang, Bo, Haizhou Ai, and Bai.
\newblock End-to-end crowd counting via joint learning local and global count.
\newblock In {\em Proc. IEEE ICIP}, pages 1215--1219, 2016.

\bibitem{simonyan2014very}
Karen Simonyan and Andrew Zisserman.
\newblock Very deep convolutional networks for large-scale image recognition.
\newblock In {\em Proc. ICLR}, 2015.

\bibitem{sindagi2017cnn}
Vishwanath~A Sindagi and Vishal~M Patel.
\newblock Cnn-based cascaded multi-task learning of high-level prior and
  density estimation for crowd counting.
\newblock In {\em Proc. IEEE AVSS}, pages 1--6, 2017.

\bibitem{sindagi2017generating}
Vishwanath~A Sindagi and Vishal~M Patel.
\newblock Generating high-quality crowd density maps using contextual pyramid
  cnns.
\newblock In {\em Proc. IEEE ICCV}, pages 1879--1888, 2017.

\bibitem{szegedy2015going}
Christian Szegedy, Wei Liu, Yangqing Jia, Pierre Sermanet, Scott Reed, Dragomir
  Anguelov, Dumitru Erhan, Vincent Vanhoucke, and Andrew Rabinovich.
\newblock Going deeper with convolutions.
\newblock In {\em Proc. IEEE CVPR}, pages 1--9, 2015.

\bibitem{viola2004robust}
Paul Viola and Michael~J Jones.
\newblock Robust real-time face detection.
\newblock {\em International Journal of Computer Vision}, 57(2):137--154, 2004.

\bibitem{walach2016learning}
Elad Walach and Lior Wolf.
\newblock Learning to count with cnn boosting.
\newblock In {\em Proc. Springer ECCV}, pages 660--676, 2016.

\bibitem{wang2004image}
Zhou Wang, Alan~C Bovik, Hamid~R Sheikh, and Eero~P Simoncelli.
\newblock Image quality assessment: from error visibility to structural
  similarity.
\newblock {\em IEEE Transactions on Image Processing}, 13(4):600--612, 2004.

\bibitem{wei2018revisiting}
Yunchao Wei, Huaxin Xiao, Honghui Shi, Zequn Jie, Jiashi Feng, and Thomas~S
  Huang.
\newblock Revisiting dilated convolution: A simple approach for weakly-and
  semi-supervised semantic segmentation.
\newblock In {\em Proc. IEEE CVPR}, pages 7268--7277, 2018.

\bibitem{zhang2015cross}
Cong Zhang, Hongsheng Li, Xiaogang Wang, and Xiaokang Yang.
\newblock Cross-scene crowd counting via deep convolutional neural networks.
\newblock In {\em Proc. IEEE CVPR}, pages 833--841, 2015.

\bibitem{zhang2017fcn}
Shanghang Zhang, Guanhang Wu, Joao~P Costeira, and Jos{\'e}~MF Moura.
\newblock Fcn-rlstm: Deep spatio-temporal neural networks for vehicle counting
  in city cameras.
\newblock In {\em Proc. IEEE ICCV}, pages 3687--3696, 2017.

\bibitem{zhang2016single}
Yingying Zhang, Desen Zhou, Siqin Chen, Shenghua Gao, and Yi Ma.
\newblock Single-image crowd counting via multi-column convolutional neural
  network.
\newblock In {\em Proc. IEEE CVPR}, pages 589--597, 2016.

\bibitem{zhou2016learning}
Bolei Zhou, Aditya Khosla, Agata Lapedriza, Aude Oliva, and Antonio Torralba.
\newblock Learning deep features for discriminative localization.
\newblock In {\em Proc. IEEE CVPR}, pages 2921--2929, 2016.

\end{thebibliography}
	}
	
\end{document}